\documentclass{article}
\usepackage{ijcai26}

\usepackage{times}
\usepackage{soul}
\usepackage{url}
\usepackage[hidelinks]{hyperref}
\usepackage[utf8]{inputenc}
\usepackage[small]{caption}
\usepackage{graphicx}
\usepackage{amsmath}
\usepackage{amsthm}
\usepackage{amssymb}
\usepackage{booktabs}
\usepackage{algorithm}
\usepackage{algorithmic}
\usepackage[switch]{lineno}

\usepackage{graphicx}
\usepackage{subcaption}
\usepackage{xcolor}
\usepackage{url}

\usepackage{tabularx,array}

\title{SirenFNO: Efficient and Full Frequency Learning of Fourier Neural Operators}

\author{
Pengqing Shi\and
Jie Yin\and
Stephen Tierney\And
Junbin Gao\\
\affiliations
The University of Sydney, Australia\\
\emails
\{pengqing.shi, jie.yin, stephen.tierney, junbin.gao\}@sydney.edu.au
}

\begin{document}

\maketitle

\begin{abstract}
    Fourier neural operators (FNOs) are effective and efficient surrogates for approximating solutions of PDEs and generalize across discretizations. However, owing to the reliance on frequency truncation to maintain learning efficiency of FNOs, empirical studies suggest that FNOs exhibit spectral bias toward low-frequency information, which may hinder the learning capability especially for certain PDEs with strong high-frequency oscillations. To address this limitation, we propose SirenFNO, a novel framework that leverages sinusoidal representation networks (SIRENs) to learn implicit neural representations and performs mode-wise kernel parameterization. Our SIREN parameterization learns a full-grid spectrum with a constant and discretization-independent parameter count, thereby eliminating the need for frequency truncation. We further extend SirenFNO with functional tensor decompositions to enhance parameter and learning efficiency. Empirical results show that our SirenFNO consistently outperforms FNO with approximately $4$ to $15$ times parameter reductions with preserved discretization invariance, and our functional decomposition variants obtain performance improvements with a maximum of $73$ times fewer parameters across multiple PDE benchmarks.
\end{abstract}

\section{Introduction}\label{sec:intro}
Operator learning aims to approximate mappings between infinite-dimensional function spaces, with applications across scientific computing and physical simulation.  %
Recently, neural operators (NOs) \cite{lu2021learning_deeponet,li2021fourier,JMLR:v24:21-1524} have been %
used to efficiently solve families of partial differential equations (PDEs), including weather forecasting \cite{pathak2022fourcastnet}, engineering design \cite{10247998}, forward/inverse problems \cite{RAISSI2018125,RAISSI2019686}.

The Fourier Neural Operator (FNO) \cite{li2021fourier} is a class of NOs that has proven highly effective and efficient for %
learning PDE solution operators. %
Specifically, FNO computes  %
the kernel integral in the frequency domain by applying the Fast Fourier Transform (FFT), performs a learned convolution there, and then maps back to the spatial domain via the inverse FFT (IFFT).  %
In practice, instead of learning the convolution weights for every frequency exhaustively, FNO performs frequency truncation with a chosen threshold to retain only low-frequency modes and discard high-frequency components. This yields %
certain computational savings, fewer parameters, and near–discretization invariance. However, the use of frequency truncation may fail to preserve the equivalence between the continuous and discrete representations of neural operators \cite{gao2025discretizationinvariance}, and hinder the model's ability to capture high-frequency details in certain PDEs with strong high-frequency oscillations \cite{xiao2024amortized}.

To address the aforementioned critical limitations that frequency truncation imposes on the FNO, we propose to parameterize the Fourier kernel via a hypernetwork. This hypernetwork performs mode-wise
kernel weight generation, dynamically producing kernel parameters for each frequency mode without any explicit truncation, thereby enabling learning across arbitrary discretization  resolutions. More importantly, as the spectral kernel learns a separate weight for each Fourier mode, the number of learnable kernel parameters in the plain FNO is determined by the modes retained after truncation, and without truncation it scales with the grid resolution.
In contrast, in our approach, the number of parameters solely depends on the hypernetwork architecture and remains fixed across different resolutions.

Specifically, we use SIREN \cite{sitzmann2019siren} %
as the hypernetwork to generate the Fourier kernel coefficients. Unlike conventional alternatives such as MLPs, which exhibit spectral bias %
toward low-frequency modes \cite{pmlr-v97-rahaman19a,qin2024betterunderstandingfourierneural,Zhi-QinJohnXuZhi-QinJohnXu2020FPFA}, SIREN's  sinusoidal activations efficiently capture %
both high- and low-frequency information. %
Therefore, SIREN provides a smooth and continuous parameterization that is agnostic to %
any specific discretization and less susceptible %
to spectral bias, making it a well-suited %
hypernetwork for the FNO  %
integral kernel. Furthermore, SIREN efficiently generates Fourier kernel coefficients with a relatively  small number of 
learnable parameters, and we further improve its efficiency by incorporating functional tensor decompositions \cite{vemuri2024FTD,vemuri2025FTD}.

\begin{figure*}[!htbp]
    \centering
    \begin{subfigure}[b]{0.48\linewidth}
        \centering
        \includegraphics[width=\linewidth]{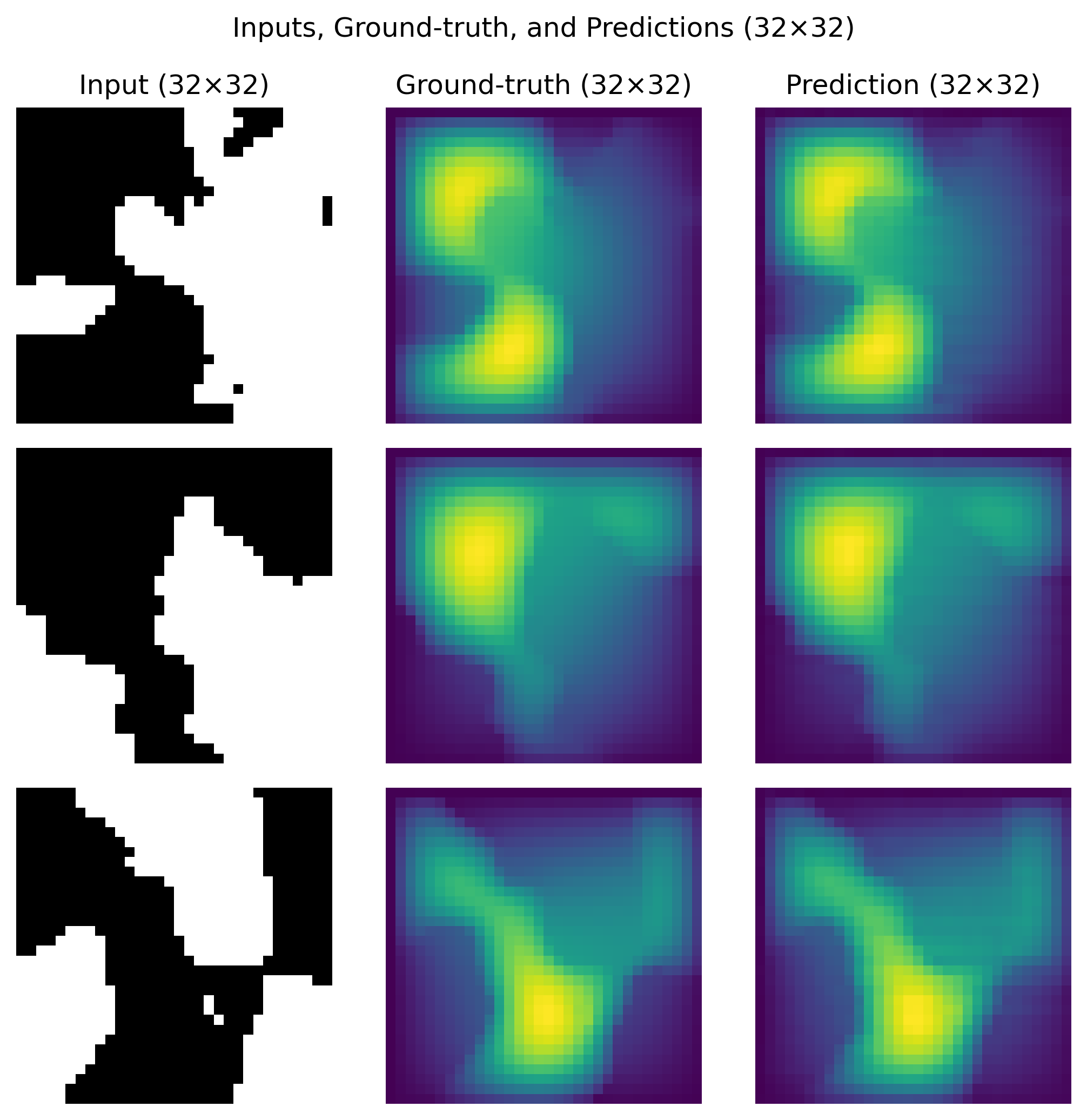}
        \caption{Trained and tested on $32 \times 32$}
        \label{fig:left}
    \end{subfigure}
    \hfill
    \begin{subfigure}[b]{0.48\linewidth}
        \centering
        \includegraphics[width=\linewidth]{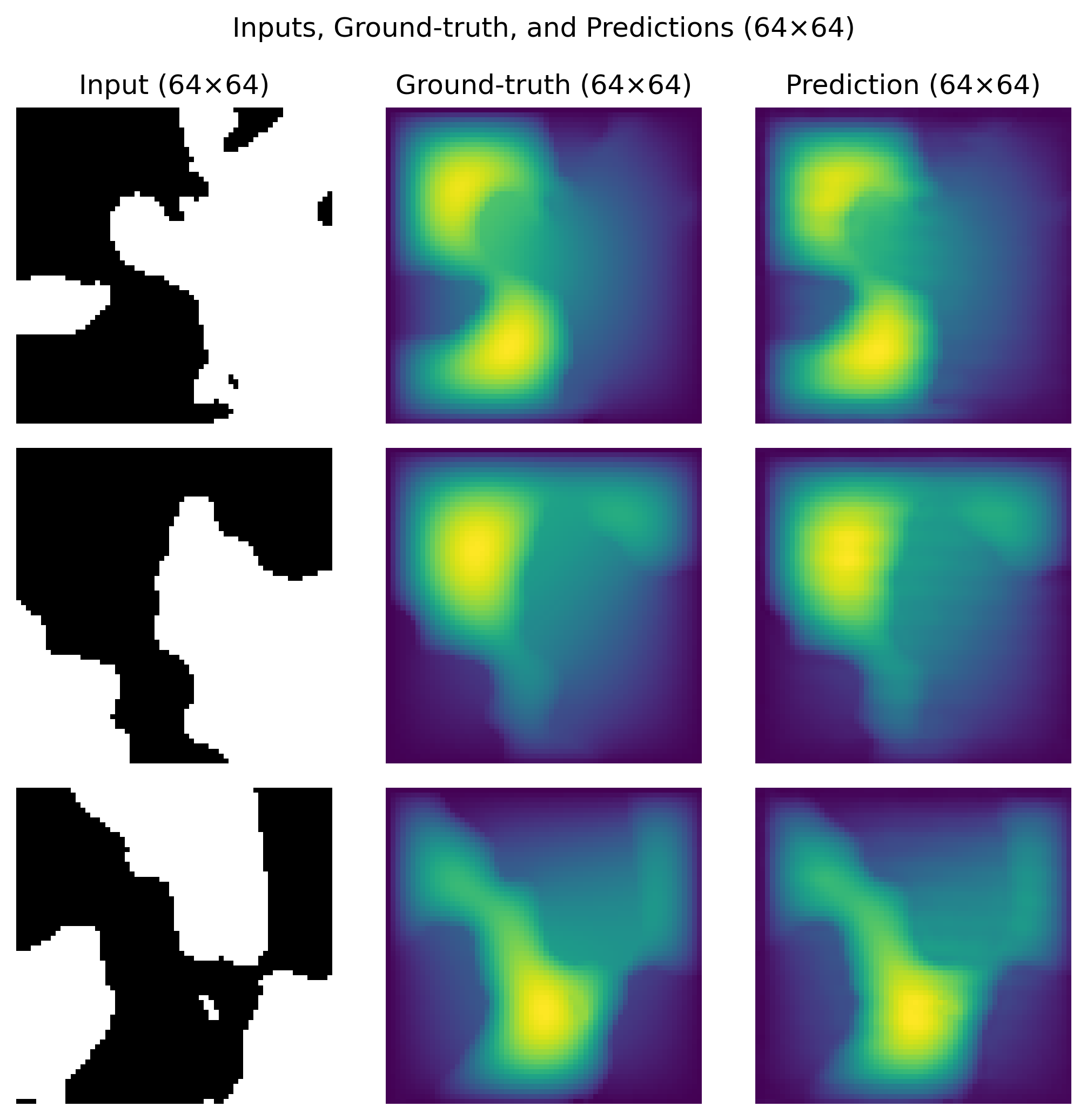}
        \caption{Trained on $32 \times 32$ and tested on $64 \times 64$}
        \label{fig:right}
    \end{subfigure}
    \caption{Zero-Shot Super-Resolution on the Darcy Flow for SirenFNO trained on $32 \times 32$ resolution.}
    \label{fig:two_subfigures}
\end{figure*}

The main contributions of this work are as follows:
\begin{itemize}
    \item We propose the Siren-FNO framework, which utilizes SIREN as a hypernetwork to generate convolution kernel weights for all frequency modes, avoiding spectral %
    truncation,  %
    while maintaining computational and parameter efficiency.
    \item We extend the Siren-FNO framework with functional tensor decompositions of the Fourier integral kernel, specifically %
    Canonical Polyadic (CP), Tensor-Train (TT), and Tucker decompositions.
    \item We demonstrate the effectiveness and resolution-agnostic behavior of the proposed methods across multiple PDE benchmarks.  %
\end{itemize}

\section{Related Works}\label{sec:related_works}

\paragraph{Neural Operators}
Neural operators learn mappings between continuous function spaces and are often used for solving PDEs. For practical implementation, these continuous functions are generally represented on discretized grids. %
DeepONet \cite{lu2021learning_deeponet} employs two sub-networks: a branch network that encodes input function values at fixed sensor points and a trunk network learns the mapping between query locations to a set of basis functions.
The outputs from these two networks are then combined to approximate operators, with guarantees from %
the universal approximation theorem. Laplace Neural Operator (LNO)
\cite{CaoQianying2024Lnof} uses the Laplace transform to decompose the input space and introduces Laplace layers to better %
learn non-periodic signals in the Laplace domain. Moreover, attention mechanisms have also been explored %
for operator learning \cite{cao2021choose}. Operator Transformer (OFormer) \cite{cao2021choose} is a transformer-based neural operator framework that employs an input encoder to embed function samples and a query encoder to project arbitrary output locations. General Neural Operator Transformer (GNOT) \cite{10.5555/3618408.3618917} introduces a heterogeneous attention operator to process diverse input function types. %
Our work focuses on the Fourier neural operators \cite{li2021fourier}.

\paragraph{Fourier Neural Operators}
The FNO \cite{li2021fourier} defines its convolution kernel in the Fourier space and uses Fast Fourier Transform (FFT) to learn the operator efficiently. Several works seek to further enhance the efficiency of the vanilla FNO. Factorized FNO (F-FNO) \cite{tran2023factorized} deepens the FNO architecture to capture long-range dependencies and factorizes layers to learn each feature dimension efficiently with fewer parameters. Adaptive FNO (AFNO) \cite{guibas2021efficient} utilizes MLP layers with learnable weights adaptively shared across frequency modes to reduce computational complexity. Beyond efficiency, motivated by the low-frequency bias of FNOs, a number of studies aim to advance its high-frequency learning ability. U-FNO \cite{wen2022u} incorporates a U-Net pathway to each Fourier layer to enhance high-frequency learning. Amortized FNO (AM-FNO) \cite{xiao2024amortized} introduces a learnable amortized parameterization of the Fourier-kernel integral that captures high-frequency information without explicit truncation. %

\begin{figure*}[!htbp]
    \centering
    \includegraphics[width=0.8\linewidth]{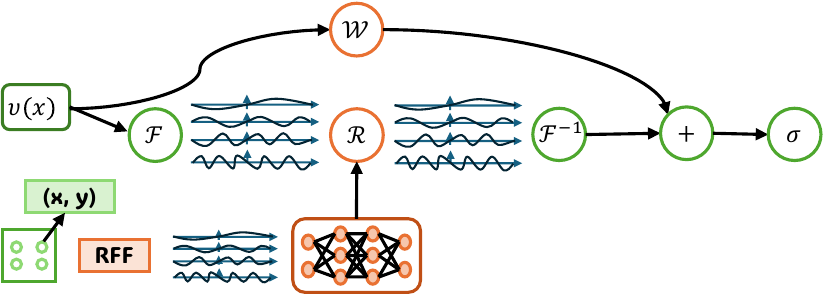}
    \caption{Illustration of SIREN Parameterization on a $2$-D PDE discretized at resolution of $2 \times 2$}
    \label{fig:siren}
\end{figure*}

\paragraph{Dynamic Hypernetworks}
Hypernetworks (hypernets) are neural networks that generate learnable weights of a primary neural network. Hypernets can be categorized by their weight-generation strategy as either %
static or dynamic %
\cite{ChauhanVinodKumar2024Abro}. A hypernet is static if its conditioning inputs and outputs have fixed dimensions for a predetermined target architecture. %
In contrast, dynamic hypernets are used to generate model weights for a primary network with a dynamic architecture, which means that the structure of the primary neural network could change or be unknown during the training and inference stages. Dynamic hypernets have been leveraged for shape reconstruction \cite{9010022}, automatic network pruning \cite{10.1007/978-3-030-58598-3_36}, or efficient architecture search \cite{PengHouwen2020Cotc}. In this work, we adopt a dynamic hypernetwork to enable parameterization generalization across discretizations.

\paragraph{Tensor Decomposition for Neural Operators}
For operator learning, tensor decomposition is often used to separate PDE dimensions or variables for improving parameter efficiency. TDMD-DeepONet \cite{CHEN2025113996} integrates tensor-train and dynamic mode decompositions into DeepONet for solving time-dependent PDEs. TensorGRaD \cite{loeschcke2025tensorgradtensorgradientrobust} directly performs Tucker decomposition on the gradient tensor. F-FNO \cite{tran2023factorized} factorizes the Fourier transforms with separable spectral layers along the problem dimensions. D-FNO \cite{LI2025117732} decomposes latent features into a sum of rank-$1$ tensor products, and substitutes the $3$-D FFT by multiple $1$-D FFTs, improving efficiency on $3$-D PDE tasks. MG-TFNO \cite{KossaifiJean2023MTFN} employs tensor decomposition in a multi-grid setting to separate FNOS Fourier convolution weights for greater parameter efficiency.

\section{Preliminaries}\label{sec:preliminaries}
This section describes the setting of the Fourier neural operator and its frequency truncation in detail.

\subsection{Fourier Neural Operators}\label{sec:fno}

For an input function space $\mathcal{A} = \mathcal{A}(D; \mathbb{R}^{d_a})$ and a target function space $\mathcal{U} = \mathcal{U}(D; \mathbb{R}^{d_u})$ on a bounded domain $D \subset \mathbb{R}^d$, operator learning aims to learn a neural operator $\mathcal{G}_{\theta}$ with parameters $\theta \in \mathbb{R}^p$ to approximate the ground-truth mapping $\mathcal{G}: \mathcal{A} \rightarrow \mathcal{U}$ from a set of observed function samples $\{a_i, u_i\}_{i=1}^N$. For a loss function $L$, we learn the neural operator by solving the problem,
\begin{equation}
    \min_{\theta} \frac1N\sum^N_{i=1}L\big(\mathcal{G}_{\theta}(a_i), u_i \big).
\end{equation}
Neural operator (NO) takes the following composition form,
\begin{equation}
    \mathcal{G}_{\theta} := \mathcal{P} \circ \mathcal{L}^{L} \circ \mathcal{L}^{L-1} \circ \cdots \circ \mathcal{L}^{1} \circ \mathcal{Q},
\end{equation}
where $\mathcal{Q}\!: \mathcal{A}(D, \mathbb{R}^{d_a}) \rightarrow \mathcal{U}(D, \mathbb{R}^{d_v})$ is the lifting operator and $\mathcal{P}\!: \mathcal{U}(D, \mathbb{R}^{d_v}) \rightarrow \mathcal{U}(D, \mathbb{R}^{d_u})$ is the projection operator, and each NO layer $\mathcal{L}^{l}$ for $l=1,2,\dots,L$ is defined as,
\begin{align}
  v^{(l+1)}&(x) = \mathcal{L}^{(l)}(v^{(l)}(\cdot))(x) := \sigma \big( W^{(l)} v^{(l)}(x) \\ &+ \int_D \kappa (x, x', a(x), a(x');\phi) v^{(l)}(x') dx' \big), \forall x \in D\notag
\end{align}
where $\sigma$ is nonlinear activation, $W \in \mathbb{R}^{d_v \times d_v}$ is a learnable weight matrix, and $\kappa$ is the integration kernel.

Fourier Neural Operator (FNO) \cite{li2021fourier} proposes a Fourier integral operator by employing convolution operation in the Fourier domain, satisfying translation invariance that $\kappa_\phi (x, x', a(x), a(x')) =\kappa_\phi (x - x')$. Applying the convolution theorem, the Fourier integral kernel is then defined as $\int_D \kappa_\phi (x - x') d x' = \mathcal{F}^{-1} ( \mathcal{F}(\kappa_\phi) \cdot \mathcal{F}(v^l))(x)$, $\forall x \in D$
Then, FNO parameterizes the periodic kernel function $\kappa$ in the Fourier domain by $\mathcal{R}_\phi(k) \in \mathbb{C}^{d_v \times d_v}$ such that $\mathcal{R}_\phi(k) = \mathcal{F}(\kappa)(k)$ for $k \in \mathbb{Z}^d$.
\begin{equation}\label{eq:FNO_layer}
    \begin{aligned}
    v^{(l+1)}(x) &= \sigma \Big( W^{(l)} v^{(l)} (x) \\ &+ \mathcal{F}^{-1} \big(\mathcal{R}_\phi^{l}(k) \cdot \mathcal{F}(v^{(l)}) (k) \big)(x) + b^{(l)} \Big), \forall x \in D
    \end{aligned}
\end{equation}

\subsection{Frequency Truncation}

\begin{figure*}[!htbp]
    \centering
    \includegraphics[width=0.9\linewidth]{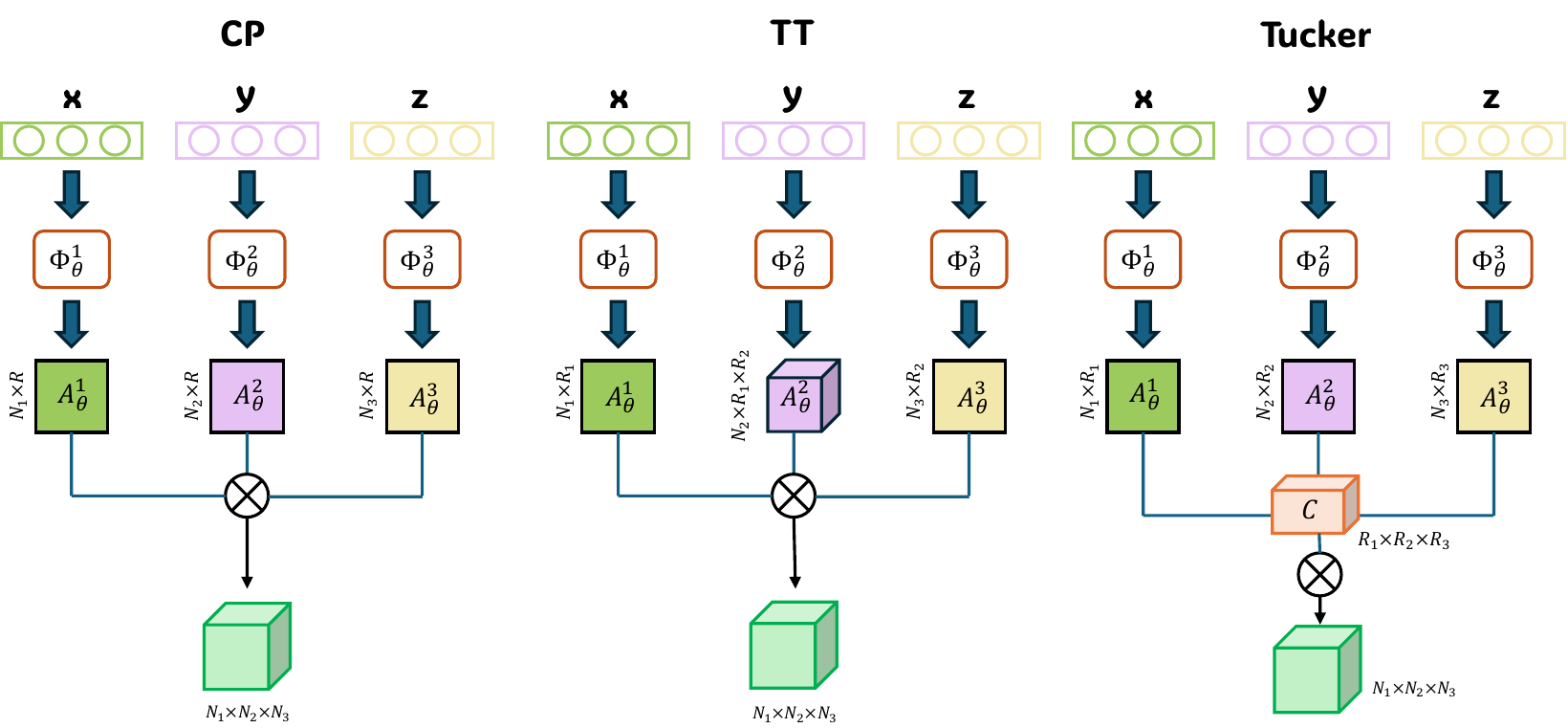}
    \caption{Example of SIREN Parameterization with Functional Tensor Decompositions on a $3$-D PDE discretized at resolution of $3 \times 3 \times 3$}
    \label{fig:siren_decomp}
\end{figure*}

Low-frequency components generally have larger magnitudes than high-frequency components in many PDE problems \cite{BoffettaGuido2012TT,george2024incremental,JING2025110912}. To prioritize the learning of low-frequency modes and improve parameter efficiency, FNO truncates the Fourier spectrum by discarding high-frequency components with a predetermined number of retained frequency modes $k_{\max} = |\{k \in \mathbb{Z}^d: |k_j|\leq k_{\max, j} \text{ for } j=1,\dots,d \}|$. In this case, we have learnable kernel parameters $R_{\phi} \in \mathbb{C}^{k_{\max} \times d_v \times d_v}$, and the size of complex-valued tensor $R_\phi$ can be greatly reduced owing to frequency truncation. Then, FNO performs zero-padding on discarded modes and applies an IFFT to recover the solution field in the physical domain.

By parameterizing only the retained low-frequency components, FNO achieves superior efficiency and captures dominant components well. However, FNO exhibits Fourier parameterization bias or spectral bias due to its inability to capture non-dominant frequencies effectively \cite{qin2024betterunderstandingfourierneural}. Moreover, the predetermined $k_{\max}$ imposes inductive bias and may degrade model performance \cite{george2024incremental}.

\section{The Proposed Method}\label{sec:proposed_method}
In this work, we propose SirenFNO, a novel framework that uses SIREN \cite{sitzmann2019siren} to generate kernel coefficients mode by mode for all discrete Fourier components, eliminating the need for truncation. %
A key advantage of SirenFNO is that the number of learnable parameters is fixed by the SIREN architecture and does not depend on %
the grid resolution. %
Thereby, our proposed SirenFNO efficiently represents %
both lower- and higher-frequency details  across varying discretizations.

\subsection{Parameterization of Fourier Kernel via SIREN}\label{sec:SirenFNO}
SIREN \cite{sitzmann2019siren} employs sinusoidal activations to %
to learn continuous implicit representations. In our parameterization, each discrete Fourier mode %
is indexed by a (normalized) frequency coordinate $\xi_k \in [-1, 1]^{d}$
for $k \in \mathbb{Z}^d$. %
A SIREN network $\Phi_\theta : [-1, 1]^{d} \rightarrow \mathbb{C}^{d_v \times d_v}$ maps $\xi_k$ to the corresponding complex Fourier kernel coefficient matrix. For a batch of modes $K$ at a given resolution, stacking inputs yields outputs in $\mathbb{C}^{|K|\times d_v \times d_v}$ the set size $|K|$ is adaptively determined by the discretization. Because all frequency coordinates are normalized to 
$[-1,1]$, the network is insensitive to changes in grid resolution. 
Specifically, for each $k$, the coefficients are produced by a SIREN stack,

\begin{equation}
    \Phi_{\theta}(\xi_k) = (\mathcal{O} \circ \phi^{(L)} \circ \cdots \circ \phi^{(1)} \circ \mathcal{E}_{B}) (\xi_k) ,\label{eq:siren_formulation}
\end{equation}
where $\mathcal{E}_{B}:\mathbb{R}^d \rightarrow \mathbb{R}^{2m}$ is a learnable random Fourier feature (RFF) embedding of the input frequency coordinates,  defined as
\begin{equation}
\mathcal{E}_{B}(\xi_k)= \big[\cos(\pi B^{\top}\xi_k),\ \sin(\pi B^{\top}\xi_k) \big] \in \mathbb{R}^{2m},
\label{eq:rff_embed}
\end{equation}
where $m$ is the number of RFF frequencies, and $B$ is a learnable parameter initialized as $B^{(0)} = \sigma G$, with $G_{ij} \stackrel{\text{i.i.d.}}{\sim} \mathcal{N}(0,1)$ and $\sigma > 0$.

Then, we have $\phi^{(1)}:\mathbb{R}^{2m} \rightarrow \mathbb{R}^h$, and $\phi^{(l)}:\mathbb{R}^{h} \rightarrow \mathbb{R}^h$ for $l = 2,\dots L$. Each SIREN layer $\phi^{(l)}$ in the above \eqref{eq:siren_formulation} takes the following form,
\begin{equation}
\xi^{(l)} \leftarrow \phi^{(l)} (\xi^{(l-1)}_k) = \sin (w \cdot W^{(l)} \xi^{(l-1)}_k + b^{(l)}),\label{eq:siren_layer}
\end{equation}
where a factor $w$ is used in each SIREN layer to accelerate convergence and span multiple periods of the sine basis, enabling our parameterization method to efficiently represent high-frequency information. Finally, $\mathcal{O}: \mathbb{R}^h \rightarrow \mathbb{C}^{d_v \times d_v}$ is an MLP layer.

Now, we replace the discreted kernel coefficients $\mathcal{R}_\phi(k) \in \mathbb{C}^{d_v \times d_v}$ in \eqref{eq:FNO_layer} with coefficients $\Phi_{\theta}(\xi_k) \in \mathbb{C}^{d_v \times d_v}$ generated by the above SIREN parameterization, our SirenFNO layer is formulated as,
\begin{align}
    v^{(l+1)}(x) &= \sigma \Big( W^{(l)} v^{(l)} (x) \label{eq:SirenFNO_layer} \\+& \mathcal{F}^{-1} \big(\Phi_{\theta}^{(l)}(\xi_k) \cdot \mathcal{F}(v^{(l)}) (k)\big)(x) + b^{(l)} \Big), \;\forall x \in D. \notag
\end{align}

In contrast to FNO, which directly maintains and optimizes a large discretization-dependent tensor of Fourier kernel coefficients, our SIREN parameterization learns a continuous mapping from frequency coordinates to kernel coefficients. This eliminates the need to store all coefficients explicitly, and they are generated on demand for each frequency mode instead. As such, SirenFNO can efficiently learn the Fourier integral operator over all frequency modes without frequency truncation, and the number of learnable parameters is independent of the discretization resolution, determined solely by the SIREN architecture. Figure~\ref{fig:siren} illustrates the process of parameter generation by our kernel parameterization.

\subsection{Functional Tensor Decomposition for SirenFNO}\label{sec:FunctionalSirenFNO}

For SirenFNO without truncation, we construct input coordinate points for SIREN block $\Phi_\theta:[-1,1]^{d}\rightarrow\mathbb{C}^{d_v\times d_v}$ on a full $d$-dimensional domain, and the SIREN block is shared across all resolution grids with $N$ points per axis.
Therefore, for a $d$-dimensional PDE discretized on a $K = N^d$ grids
, the construction of input coordinates for SIREN takes up $O(N^d \times d)$ memory, which requires excessive memory usage, especially for finer grids and high-dimensional PDE applications. To alleviate this, we further extend  SirenFNO by leveraging functional tensor decompositions.

In this work, we employ three tensor decomposition schemes, %
\textit{Canonic-Polyadic (CP)}, \textit{Tensor-Train (TT)}, and \textit{Tucker}, within SirenFNO. %
In the functional tensor variants, the kernel is parameterized  by factorizing the SIREN trunk into $d$ branch sub-networks, one per %
geometric dimension.  %
Instead of sampling a full $N^d$ grid, %
each branch processes $N$ coordinates for its dimension, yielding a factor vector/matrix; the full tensor of Fourier-kernel coefficients is then reconstructed via tensor (outer) products. %
This reduces the sampling requirement from $N^d$ points to $N \times d$ and the memory footprint %
from exponential to linear in $d$, i.e. $\mathcal{O}(Nd)$. %

\paragraph{CP-SirenFNO.}
With CP rank \(R \ll \min_{i \in [d]} N_i\), the kernel tensor \(\Phi_{\theta}\in\mathbb{C}^{N_1\times\cdots\times N_d}\) is  parameterized as a sum of \(R\) rank-1 outer products. SirenFNO realizes this by using \(d\) SIREN branches to produce factor matrices \(A^{(i)}\in\mathbb{R}^{N_i\times R}\) for \(i=1,\ldots,d\). Let \(a^{(i)}_{r}:=A^{(i)}(:,r)\in\mathbb{R}^{N_i}\) denote the \(r\)-th column. The reconstruction is
\begin{equation}
\Phi_{\theta} = \sum_{r=1}^{R} a^{(1)}_{r} \otimes a^{(2)}_{r}\ \circ\ \cdots\ \circ\ a^{(d)}_{r},
\label{eq:CP_decomp}
\end{equation}
where \(\otimes\) denotes the outer (tensor) product. This reduces storage from \(\prod_{i=1}^{d} N_i\) to \(\mathcal{O}\!\big(R\sum_{i=1}^{d} N_i\big)\) while enabling on-demand coefficient generation.

\paragraph{TT-SirenFNO.}
With TT ranks \(R_0=R_d=1\) and \({R_1,\ldots,R_{d-1}}\), SirenFNO uses \(d\) SIREN branches to produce cores \(G^{(i)}\in\mathbb{C}^{R_{i-1}\times N_i \times R_i}\) for \(i=1,\ldots,d\). Let \(g^{(i)}_{r_{i-1},r_i}\in\mathbb{C}^{N_i}\) denote the mode-2 fiber of \(G^{(i)}\) with TT indices \((r_{i-1},r_i)\) fixed. The kernel tensor \(\Phi_{\theta}\in\mathbb{C}^{N_1\times\cdots\times N_d}\) is reconstructed as
\begin{equation}
\Phi_{\theta} = \sum_{r_1=1}^{R_1}\cdots\sum_{r_{d-1}=1}^{R_{d-1}}
g^{(1)}_{1,r_1}\ \otimes \ g^{(2)}_{r_1,r_2}\ \otimes\ \cdots\ \otimes g^{(d)}_{r_{d-1},1}.
\label{eq:TT_decomp}
\end{equation}
This representation reduces storage from \(\prod_{i=1}^{d}N_i\) to \(\mathcal{O}\big(\sum_{i=1}^{d} N_i R_{i-1}R_i\big)\).

\paragraph{Tucker-SirenFNO.}
Given Tucker ranks \({R_1,\ldots,R_d}\), SirenFNO employs \(d\) SIREN branches to produce factor matrices \(A^{(i)}\in\mathbb{C}^{N_i\times R_i}\) for \(i=1,\ldots,d\), together with a learnable core \(\mathcal{C}\in\mathbb{C}^{R_1\times\cdots\times R_d}\). The kernel tensor \(\Phi_{\theta}\in\mathbb{C}^{N_1\times\cdots\times N_d}\) is reconstructed via the Tucker product
\begin{equation}
\Phi_{\theta}= \mathcal{C} \times_1 A^{(1)} \times_2 A^{(2)} \cdots \times_d A^{(d)},
\label{eq:Tucker_nmode}
\end{equation}
where $\times_i$ means mode-$i$ tensor-matrix product.

For SirenFNO with functional tensor decompositions, these factor matrices $\{A^1,\dots,A^d\}$ are generated and parameterized using a number of $d$ small SIREN networks $\{\Phi^1_{\theta_1},\dots,\Phi^d_{\theta_d}\}$ accordingly. The reconstructed tensor is then used as Fourier kernel coefficients in the FNO. Figure~\ref{fig:siren_decomp} illustrates an example of SIREN parameterization with functional tensor decompositions.

\subsection{Model Architecture}
Our SirenFNO follows the standard FNO architecture,
\begin{equation}
    \mathcal{G}_{\theta} := \mathcal{P} \circ \mathcal{L}^{L} \circ \mathcal{L}^{L-1} \circ \cdots \circ \mathcal{L}^{1} \circ \mathcal{Q},
\end{equation}
where $\mathcal{Q}$ is the lifting layer to transform input data $\alpha$ to hidden state $z$, and $\mathcal{P}$ is the projection layer to project hidden state $z$ to output prediction $u$. Furthermore, intermediary layers $\mathcal{L}^{l}$ for $l=1, \dots, L$ are our SirenFNO layers introduced in \eqref{eq:SirenFNO_layer}. Now, we use $\mathcal{K}_\Phi (v^{(l)}(z)) := \mathcal{F}^{-1} \big(\Phi_{\theta}^{(l)}(\xi_k) \cdot \mathcal{F}(v^{(l)}) (k)\big)(z)$ to denote our SirenFNO parameterization kernel, we have
\begin{equation}
    v^{(l+1)}(z) = \sigma \Big( W^{(l)} v^{(l)} (z) + \mathcal{K}_\Phi (v^{(l)}(z)) + b^{(l)} \Big), \quad \forall z \in D.
\end{equation}
In this work, we leverage the same model architecture as AM-FNO \cite{xiao2024amortized,tran2023factorized}, using residual connections and replacing linear mapping $W^l v^l (z)$ such that

\begin{align}
    v^{(l+1)}&(z) = \sigma \Big( v^{(l)} (z) + \\& W^{(l)}_2  \sigma ( W^{(l)}_1 \mathcal{K}_\Phi (v^{(l)}(z)) + b_1^{(l)}) +b_2^{(l)} \Big), \quad \forall z \in D. \notag
\end{align}
It is worth noting that each SirenFNO operator layer includes additional nonlinearities after the residual connection, and the last operator layer omits the activation after the residual connection.

\section{Experiments}\label{sec:experiments}

This section describes our experimental setups and results. All experiments are conducted on the same machine equipped with an NVIDIA GeForce RTX 5090. Our Code is available at \url{https://github.com/pengqingshi/SirenFNO}.

\subsection{Benchmarks and Baselines}

We compare our proposed models against FNO \cite{li2021fourier} and its several well-known variants, including UFNO in \cite{wen2022u}, Tensorized FNO with CP decomposition (TFNO-CP) \cite{KossaifiJean2023MTFN}, and AM-FNO with MLPs parameterization \cite{xiao2024amortized}, and UNO in \cite{rahman2023uno}.

\newcolumntype{Y}{>{\raggedright\arraybackslash}X}
\begin{table}[H]
  \centering
  \begingroup
  \scriptsize
  \setlength{\tabcolsep}{3.5pt}
  \begin{tabularx}{\columnwidth}{@{}l Y c c c c@{}}
    \hline
    Name & Dataset & Resolution & No. train & No. test \\
    \hline
    Darcy    & 2D Darcy flow                     & $128\!\times\!128$ & $1000$ & $200$ \\
    NS       & 2D incompressible Navier Stokes   & $128\!\times\!128$                 & $1000$ & $200$ \\
    Burgers  & 1D Burgers’ equation              & $1024, T\!=\!20$                   & $1000$ & $200$ \\
    1DCFD    & 1D computational fluid dynamics   & $1024, T\!=\!20$                   & $1800$ & $200$ \\
    2DCFD    & 2D computational fluid dynamics   & $128\!\times\!128, T\!=\!10$       & $1800$ & $200$ \\
    ReacDiff & 1D Reaction-Diffusion             & $1024, T\!=\!20$                   & $1000$ & $200$ \\
    \hline
  \end{tabularx}
  \endgroup
  \caption{Description of benchmark datasets.}
  \label{tab:datasets}
\end{table}

\begin{table*}[t]
\centering
\begin{tabular}{|l|c|c|c|c|c|c|c|}
\hline
Dataset & \multicolumn{2}{|c|}{\textbf{Darcy}} & \textbf{NS} & \textbf{Burgers} & \textbf{1DCFD} & \textbf{2DCFD} & \textbf{ReacDiff}\\
\hline
Resolution & $32 \times 32$ & $128 \times 128$ & $128 \times 128$ & $1024$ & $1024$ & $128 \times 128$ & $1024$\\
\hline
FNO & $7.30e$-$2$ & $6.26e$-$2$ & $5.61e$-$2$ & $1.12e$-$2$ & $8.27e$-$3$ & $1.64e$-$2$ & $1.18e$-$4$\\
UFNO & $7.07e$-$2$ & $6.00e$-$2$ & $5.37e$-$2$ & $1.70e$-$2$ & $8.29e$-$3$ & $1.21e$-$2$ & $1.98e$-$4$\\
TFNO-CP & $5.20e$-$2$ & $4.81e$-$2$ & $5.17e$-$2$ & $1.05e$-$2$ & $6.31e$-$3$ & $8.25e$-$3$ & $1.14e$-$4$\\
U-NO & $6.57e$-$2$ & $7.68e$-$2$ & $4.96e$-$2$ & $7.53e$-$3$ & $7.62e$-$3$ & $1.62e$-$2$ & $7.79e$-$5$\\
AM-FNO (MLP) & $3.72e$-$2$ & $5.10e$-$2$ & $4.02e$-$2$ & $8.92e$-$3$ & $7.89e$-$3$ & $1.21e$-$2$ & $1.78e$-$3$\\
\hline
SirenFNO\textsuperscript{\dag} & $6.32e$-$2$ & $5.82e$-$2$ & $5.52e$-$2$ & $8.21e$-$3$ & $7.41e$-$3$ & $1.15e$-$2$ & $7.40e$-$5$\\
CP-SirenFNO\textsuperscript{\dag} & $4.99e$-$2$ & $3.96e$-$2$ & $5.04e$-$2$ & $8.12e$-$3$ & $5.91e$-$3$ & $7.28e$-$3$ & $8.98e$-$5$\\
TT-SirenFNO\textsuperscript{\dag} & $5.42e$-$2$ & $5.36e$-$2$ & $5.87e$-$2$ & $7.94e$-$3$ & $5.58e$-$3$ & $8.75e$-$3$ & $8.51e$-$5$\\
Tucker-SirenFNO\textsuperscript{\dag} & $4.45e$-$2$ & $4.51e$-$2$ & $4.19e$-$2$ & $7.18e$-$3$ & $7.31e$-$3$ & $7.30e$-$3$ & $7.81e$-$5$\\
\hline
\textbf{SirenFNO} & $\mathbf{3.51\textbf{e-}2}$ & $\mathbf{2.36\textbf{e-}2}$ & $3.51e$-$2$ & $\mathbf{5.52\textbf{e-}3}$ & $6.77e$-$3$ & $\mathbf{6.83\textbf{e-}3}$ & $\mathbf{6.31\textbf{e-}5}$\\
\textbf{CP-SirenFNO} & $4.04e$-$2$ & $3.03e$-$2$ & $3.78e$-$2$ & $8.48e$-$3$ & $\mathbf{4.75\textbf{e-}3}$ & $7.40e$-$3$ & $7.58e$-$5$\\
\textbf{TT-SirenFNO} & $4.18e$-$2$ & $3.33e$-$2$ & $\mathbf{3.24\textbf{e-}2}$ & $7.52e$-$3$ & $6.32e$-$3$ & $7.41e$-$3$ & $8.92e$-$5$\\
\textbf{Tucker-SirenFNO} & $3.98e$-$2$ & $3.18e$-$2$ & $3.86e$-$2$ & $8.11e$-$3$ & $5.29e$-$3$ & $6.90e$-$3$ & $8.71e$-$5$\\
\hline
\end{tabular}
\caption{$\ell_2$ relative test errors. All models are train and test on the same corresponding resolution. Models highlighted with \textsuperscript{\dag} are using the identical architecture with FNO for ablation study. SirenFNO and its variants in boldface but without \textsuperscript{\dag} are our proposed methods. Note that our proposed methods learn full-frequency information without truncation.}
\label{tab:exp_main_results}
\end{table*}

Table~\ref{tab:datasets} lists the benchmark datasets used in our experiments. Darcy and Navier-Stokes datasets are obtained from the NeuralOperator library \cite{kossaifi2024neural,JMLR:v24:21-1524}; and 1D Burgers\footnote{\path{1D_Burgers_Sols_Nu0.001.hdf5}}, 1D CFD\footnote{\path{1D_CFD_Rand_Eta0.01_Zeta0.01_periodic_Train.hdf5}}, 2D CFD\footnote{\path{2D_CFD_Rand_M0.1_Eta0.01_Zeta0.01_periodic_128_Train.hdf5}}, and reaction-diffusion\footnote{\path{ReacDiff_Nu0.5_Rho1.0.hdf5}} are sourced from PDEBench \cite{pdebench_Makoto2024}.

\enlargethispage{-2\baselineskip}

\subsection{Implementation Details}
The $\ell_2$ relative error is utilized for training and evaluation. Given data pairs $\{a_i, u_i \}_{i=1}^N$ and a model $\mathcal{G}_\theta: \mathcal{A} \rightarrow \mathcal{U}$, the $\ell_2$ relative error is computed as,
\begin{equation}
    L (\mathcal{G}_\theta(a), u) = \frac1N \sum_{i=1}^N \frac{\|\mathcal{G}_\theta(a_i) - u_i \|^2}{\| u_i \|^2}.
\end{equation}

For each experiment, we conduct $500$ training epochs using AdamW optimizer with a batch size of $32$, a learning rate of $1e\text{-}3$ with the cosine annealing schedule and a weight decay of $1e\text{-}4$. We set $4$ layers with $32$ hidden channels for FNO and its variants by default, including our SirenFNO. The hidden channels for lifting and projection layers are set to $64$ by default. For FNO frequency truncation, we retain modes of $m_x=m_y=32$ for Navier-Stokes and 2D CFD, and $m_x=m_y=16$ for the Darcy flow. We learn full frequency for FNO on Burgers, 1D CFD, and reaction-diffusion. Please note that our proposed SirenFNO and its variant forms do not perform frequency truncation. Additionally, We evaluate time-dependent PDEs in an one-step autoregressive scheme, where we perform $10$ rollout steps with previous $10$ states for Burgers, reaction-diffusion, and 1D CFD; and we conduct $5$ rollout steps with previous $5$ states for 2D CFD.

\subsection{Main Results}

\begin{table*}[!htbp]
\centering
\begin{tabular}{|l|c|c|c|c|c|c|c|}
\hline
Dataset & \multicolumn{2}{|c|}{\textbf{Darcy}} & \textbf{NS} & \textbf{Burgers} & \textbf{1DCFD} & \textbf{2DCFD} & \textbf{ReacDiff}\\
\hline
Resolution & $32 \times 32$ & $128 \times 128$ & $128 \times 128$ & $1024$ & $1024$ & $128 \times 128$ & $1024$\\
\hline
FNO & $1192.8$ & $1192.8$ & $4469.6$ & $4216.2$ & $4216.2$ & $4469.9$ & $4216.2$\\
UFNO & $5876.2$ & $5876.2$ & $990.8$ & $4258.5$ & $1892.2$ & $991.0$ & $1892.2$\\
TFNO-CP & $72.9$ & $72.9$ & $237.5$ & $226.4$ & $226.4$ & $237.8$ & $226.4$\\
U-NO & $1792.7$ & $4250.3$ & $8726.8$ & $1726.1$ & $1726.1$ & $7602.3$ & $5658.2$\\
AM-FNO (MLP) & $385.5$ & $385.5$ & $4443.1$ & $207.6$ & $823.1$ & $385.6$ & $823.1$\\
\hline
\textbf{SirenFNO} & $308.9$ & $308.9$ & $579.5$ & $308.9$ & $304.8$ & $304.8$ & $304.8$\\
\textbf{CP-SirenFNO} & $\mathbf{63.9}$ & $\mathbf{63.9}$ & $\textbf{138.9}$ & $\mathbf{70.1}$ & $\mathbf{57.7}$ & $\mathbf{64.0}$ & $\textbf{57.7}$\\
\textbf{TT-SirenFNO} & $92.5$ & $109.2$ & $211.6$ & $84.5$ & $72.0$ & $92.7$ & $72.0$\\
\textbf{Tucker-SirenFNO} & $162.6$ & $162.6$ & $596.4$ & $74.2$ & $61.8$ & $96.8$ & $61.8$\\
\hline
\end{tabular}
\caption{Number of parameters used for each experiment, numbers are reported in thousands approximately. Note that our proposed methods learn full-frequency information without truncation.}
\label{tab:exp_main_results_params}
\end{table*}

Table~\ref{tab:exp_main_results} reports the $\ell_2$ relative test set error for each PDE benchmark, with all models trained and tested at identical discretization resolutions. Table~\ref{tab:exp_main_results_params} lists the number of learnable parameters in thousands for every neural operator in each experiment. For each problem, we boldface the lowest $\ell_2$ relative test error and the smallest parameter count.

As shown in Table~\ref{tab:exp_main_results}, our SirenFNO and its functional decomposition variants consistently surpass the vanilla FNO across all PDE benchmarks. Moreover, as indicated in Table~\ref{tab:exp_main_results_params}, they achieve remarkably greater parameter efficiency without relying on frequency truncation.

For the $2$D Darcy flow problem, our SirenFNO reduces the $\ell_2$ relative test error from $0.0730$ to $0.0351$ (by $52\%$) and from $0.0626$ to $0.0236$ (by $62\%$) for discretizations of $32 \times 32$ and $128 \times 128$ respectively, with an approximately $4\times$ reduction in model parameters, compared to FNO. Moreover, CP, TT, and Tucker variants maintain comparable performance while further reducing the number of parameters. For solving the $2$D Navier-Stokes equation, our SirenFNO obtains a $\ell_2$ relative error of $0.0351$, compared to that of $0.0561$ for FNO; while using about $8$ times fewer parameters. 

Notably, our TT-SirenFNO achieves the best overall performance on the Navier-Stokes benchmark, with a further reduction in parameters by $21$ times compared to FNO. On the $1$D Burgers benchmark, SirenFNO yields the lowest error, which is $51\%$ lower than that of FNO, with about $14$ times fewer parameters. CP‑SirenFNO also outperforms FNO with roughly $60$ times fewer parameters, and the TT and Tucker variants similarly improve over FNO with about $50$ and $57$ times fewer parameters. For 1D CFD, SirenFNO reduces the $\ell_2$ error by $18\%$ lower) while using $14$ times fewer parameters than FNO, and CP-SirenFNO is the best-performing model with only $57.7$ thousands parameters. Similarly, for 2D CFD and reaction-diffusion, SirenFNO performs the best among all, while using significantly lower parameters than standard FNO. The improvement gains can be attributed to the ability of SIREN parameterization to capture high- and low-frequency information, enabling full-frequency learning irrespective of discretizations.

\subsection{Ablation Study}

As shown in Table~\ref{tab:exp_main_results}, our proposed methods adopt the enhanced architecture used in AM-FNO (MLP). SirenFNO consistently outperforms all baselines while using significantly fewer parameters to learn full-frequency information, demonstrating both the effectiveness and efficiency of our model.

For the ablation study, we further isolate the effect of architectural changes by keeping the operator architecture identical to the standard FNO implementation. According to Table~\ref{tab:exp_main_results}, under a strictly controlled setting, SirenFNO\textsuperscript{\dag} results indicate that our SIREN parameterization yields consistent performance gains across all PDE benchmarks. Performance gains can be purely attributed to our kernel parameterization, suggesting that our methods mitigate the spectral bias of FNO and enhance its high-frequency learning capability while achieving superior parameter efficiency. Furthermore, our SirenFNO delivers further improvements, indicating a complementary and synergic effect by employing the improved operator architecture.

\subsection{Zero-Shot Super-Resolution}

A key advantage of FNO is resolution invariance, which enables a model trained on a coarse grid to generalize well to a finer grid without retraining. To preserve resolution-invariant generalization for our SirenFNO, we adopt frequency truncation with modes $m_x=m_y=12$. Table~\ref{tab:zero-shot} reports the zero-shot super-resolution performance on $2$D Darcy flow, and Figure~\ref{fig:two_subfigures} visualizes the predictions. All models are trained on a discretization of $32 \times 32$ and then evaluated on $32 \times 32$ and $64 \times 64$ grids. Our SirenFNO consistently outperforms FNO and maintains strong performance across resolutions.

\begin{table}[!htbp]
    \centering
    \begin{tabular}{|l|c|c|}
    \hline
     & $32\times 32$ & $64 \times 64$\\
     \hline
    FNO & $0.0730$ & $0.0803$\\
    SirenFNO & $0.0451$ & $0.0599$\\
    CP-SirenFNO & $0.0502$ & $0.0606$\\
    TT-SirenFNO & $0.0459$ & $0.0591$\\
    Tucker-SirenFNO & $\textbf{0.0455}$ & $\textbf{0.0576}$\\
    \hline
    \end{tabular}
    \caption{$\ell_2$ relative test error for zero-shot super-resolution performance on the $2$D Darcy flow. Models are trained on $32 \times 32$ resolution.}
    \label{tab:zero-shot}
\end{table}

\section{Conclusion}
In this work, we propose SirenFNO to perform full-frequency learning and mitigate spectral bias without truncating Fourier modes, we also consider functional tensor decompositions with CP, TT, and Tucker variants to further improve the parameter and learning efficiency. As suggested by our experiments across various PDE benchmarks, our SirenFNO and its variants surpass FNO in terms of $\ell_2$ relative test error, using approximately $4$ to $73$ times fewer parameters without truncation. Overall, our proposed SirenFNO provides a more accurate and efficient alternative to standard FNO architectures for operator learning on PDE benchmarks.

\bibliographystyle{named}
\bibliography{ijcai26}

\end{document}